\documentclass[letterpaper, 10pt, conference]{IEEEtran}

\IEEEoverridecommandlockouts                              

\usepackage{amsmath} 
\usepackage{xcolor}
\usepackage{graphicx}
\usepackage{tabularx}
\usepackage{hyperref}
\usepackage{cleveref}
\usepackage{caption}
\usepackage{cuted}
\usepackage{makecell}
\usepackage{amsfonts}
\usepackage{bm}
\usepackage[numbers]{natbib}
\usepackage[acronym,shortcuts]{glossaries}

\usepackage{eso-pic}

\newcommand\AtPageUpperCenterNotice[1]{%
  \AtPageUpperLeft{%
    \put(\LenToUnit{0.5\paperwidth},\LenToUnit{-2cm}){\makebox[0pt]{#1}}%
  }%
}

\AddToShipoutPictureBG*{%
  \AtPageUpperCenterNotice{%
    \parbox[b][0.5cm][c]{\paperwidth}{%
      \centering
      \fontsize{12}{14}\selectfont
      \color{gray!50}
      This paper has been accepted for publication in\\
       International Conference on Robotics and Automation (ICRA), Vienna 2026. \copyright{}IEEE.
    }%
  }%
}

\AddToShipoutPictureBG*{
  \AtPageLowerLeft{%
    \raisebox{25pt}{\makebox[\paperwidth]{\begin{minipage}{21cm}\centering
    \fontsize{10}{8}\selectfont
    \textcolor{gray!50}{%
      \copyright{} 2026 IEEE. Personal use of this material is permitted.
      Permission from IEEE must be obtained for all other uses, in any current or future media,
      including reprinting/republishing this material for advertising or promotional purposes,
      creating new collective works, for resale or redistribution to servers or lists,
      or reuse of any copyrighted component of this work in other works.
    }
    \end{minipage}}}%
  }
}

\title{\LARGE \bf
\vspace{6.3mm}
Continual-RL for Generalization in Autonomous Racing on the RoboRacer Platform
}

\author{Joel Siegert$^{1,\dagger}$, Edoardo Ghignone$^{1,\dagger}$, Michele Magno$^{1}$
\thanks{*This work was not supported by any organization}
\thanks{$^{1}$Center for Project-Based Learning, D-ITET, ETH Zurich,
        {\tt\small edoardo.ghignone@pbl.ee.ethz.ch}}%
\thanks{$^\dagger$Equal Contribution}
}

\begin{document}
\bstctlcite{IEEEexample:BSTcontrol}

\newacronym{rl}{RL}{Reinforcement Learning}
\newacronym{sota}{SotA}{State-of-the-Art}
\newacronym{ar}{AR}{Autonomous Racing}
\newacronym{cbp}{CBP}{Continual Backpropagation}
\newacronym{iql}{IQL}{Implicit Q-Learning}
\newacronym{e2e}{E2E}{End-to-End}
\newacronym{sde}{SDE}{State-Dependent Exploration}
\newacronym{ere}{ERE}{Emphasizing Recent Experiences}
\newacronym{redo}{ReDo}{Reinitialize Dormant Neurons}
\newacronym{s2e}{S2E}{State-to-End}
\newacronym{lidar}{LiDAR}{Light Detection and Ranging}
\newacronym{imu}{IMU}{Inertial Measurement Unit}
\newacronym{sb3}{SB3}{Stable Baselines 3}
\newacronym{ros}{ROS}{Robot Operating System}
\newacronym{gsde}{gSDE}{Generalized State-Dependent Exploration}
\newacronym{erpm}{ERPM}{Electrical Revolutions Per Minute}
\newacronym{sd}{SD}{Standard Deviation}
\newacronym{mpcc}{MPCC}{Model Predictive Contouring Control}
\newacronym{mpc}{MPC}{Model Predictive Control}
\newacronym{utd}{UTD}{Update-To-Data}
\newacronym{knn}{k-NN}{k-Nearest Neighbors}
\newacronym{pp}{PP}{Pure Pursuit}

\def\methoda{CBP-SAC}
\def\methodb{IQL-CBP-SAC}

\newacronym{pp}{PP}{Pure Pursuit}
\newacronym{map}{MAP}{Model- and Acceleration-based Pursuit}
\newacronym{ml}{ML}{Machine Learning}
\newacronym{nn}{NN}{Neural Network}
\newacronym{vesc}{VESC}{Vedder Electronic Speed Controller}
\newacronym{cots}{CotS}{Commercial off-the-Shelf}
\newacronym{sac}{SAC}{Soft Actor-Critic}
\newacronym{mlp}{MLP}{Multilayer Perceptron}
\newacronym{lut}{LuT}{Lookup Table}

\maketitle
\thispagestyle{empty}
\pagestyle{empty}

\begin{abstract}
A key challenge in modern robotics is to adapt to changing environments, a challenge that is exacerbated when simulations cannot encompass every possible real-world configuration, and therefore \acrfull{rl} in the physical world becomes necessary.
Continual \gls{rl} provides the tools to address this challenge; however, both the frameworks and the methods remain underexplored.
\acrfull{ar} and in particular the RoboRacer competition provide a testing ground for such methods, as learning to drive on a new track-floor combination with the least amount of new experience naturally frames a continual learning problem.
This work tries to address this gap by proposing a continual \gls{rl} framework based on \acrfull{cbp} that is able, with only real-world data, to train a generalistic policy on a set of tracks and then fine-tune it within 15 minutes to outperform classical controllers. Furthermore, a comparison method based on offline \gls{rl} is proposed, and a simulation analysis of the plasticity properties of the methods is conducted.
\end{abstract}

\section{INTRODUCTION}
\gls{rl} has recently evolved into a practical tool for robotics, achieving superhuman performance in board games \cite{alpha_go}, quadrupeds \cite{circus_anymal, go2_quadruped}, and drone racing \cite{drone_racing}.
Generalization to the physical world remains a central challenge, usually tackled by carefully designing the simulator used for pre-training and adding randomization on key parameters: some examples are legged locomotion on unknown terrain \cite{anymal_terrain} or grasping novel objects \cite{dexpoint}.

Following a different approach, this paper focuses on direct real-world training, as simulators can be expensive or difficult to set up \cite{simulators_study}, and generalization via randomization can further add over-conservativeness.
Real-world training, though, only allows training of one agent in a single environment at a time (given one robot), and this drastically reduces the amount of data available, highlighting the need for sample-efficient algorithms.
Furthermore, the real world might present non-stationary environments, which introduce continual learning challenges: maintaining adaptability without catastrophic forgetting is a requirement \cite{computationally_constrained_rl}.


\gls{ar} on the RoboRacer platform (formerly F1TENTH) provides an affordable yet demanding testbed for benchmarking \gls{sota} controllers on real hardware in non-stationary environments: firstly, the ever-changing track layout highlights the need for algorithms that can adapt to different settings, such as continual learning methods; furthermore, the tight schedule of the race imposes a strict constraint on the available samples. Algorithms have to adapt with only minutes and not hours or days of extra data.
Different previous \gls{rl} methods have been applied to the RoboRacer platform \cite{e2e_lidar_driven_rl, fastrlap, tiny_lidar_net, hildisch2025drive}, but no result has been a viable solution able to generalize across different real-world track layouts.

In this context, we present a continual RoboRacer training framework that adopts the off-policy \gls{sac} architecture \cite{haarnoja2018sac}, used for its renowned sample efficiency, and adapt it to a continual setup with \gls{cbp} \cite{cbp2024}.

\begin{figure*}[t]
    \centering
    \includegraphics[width=0.98\textwidth]{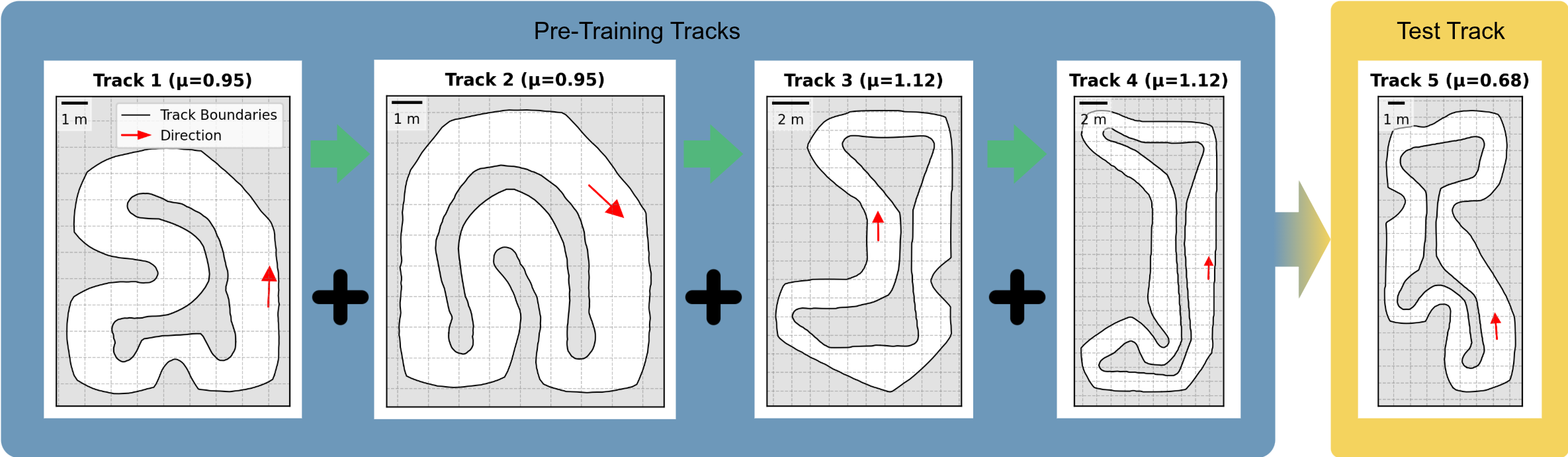}
    \caption{Set of tracks used for this study, subdivided into four pre-training and one test track. A Newton-meter was used to measure lateral static friction (indicated with $\mu$). The \textcolor[HTML]{52B77C}{\textbf{green}} arrows indicate the sequential order in which the tracks are used for continual training, while the \textbf{black} addition signs indicate the alternative offline pre-training strategy.}
    \label{fig:track_setup}
\end{figure*}

In particular, we conduct tests assuming multiple tracks can be available for pre-training, but a final track is unknown (see \Cref{fig:track_setup} for more details), and we show that the combination of \gls{cbp} with \gls{sac} can surpass one of the leading controllers in the RoboRacer competition within 15 minutes of fine-tuning.
Furthermore, we compare this first method to offline \gls{rl} pre-training by means of \gls{iql} \cite{kostrikov_offline_2021}, showing that even though offline-RL can train a model on the entire dataset, continual RL can adapt quicker to an unseen track layout and tire-floor combination.

Our summarized contributions are:
\begin{itemize}
    \item A continual \gls{rl} framework based on \gls{cbp} that learns generalistic policies with only real-world data and fine-tunes within 15 minutes on an unseen track-floor combination. Previous work showed similar performance after only 82 minutes \cite{hildisch2025drive}.
    \item An alternative framework for generalistic policy pre-training based on offline \gls{rl}, that shows promising signs of plasticity but reduced performance, at parity of training steps. 
    \item An evaluation in simulation of the effect of adding continual learning network modifications (\gls{cbp} \cite{cbp2024}, L2 Init \cite{l2_init}) to only buffer management strategies, showing that continual techniques help the performance increase from a fine-tuning stage.
    \item Tracks, simulation models, and \gls{rl} frameworks are open-sourced to foster replication and comparisons (\href{github.com/ForzaETH/Continual-RL-ICRA-26}{github.com/ForzaETH/Continual-RL-ICRA-26})
\end{itemize}

\section{RELATED WORK}
\subsection{Real-World \acrlong{rl}}
Most of the success of \gls{rl} in the real world comes from policies pretrained in simulation \cite{rudin2025parkourwildlearninggeneral, drone_racing, DBLP:journals/nature/DegraveFBNTCEHA22}. However, the sim-to-real gap is a remaining challenge: most successful transfers require at least some amount of real data or fine-tuning on the real system \cite{annurev2024DRLrealworld, djeumou2024referencefreeformuladriftreinforcement}, zero-shot transfer can often be impractical due to the sim-to-real gap \cite{annurev2024DRLrealworld, sim2real_problem}, and domain randomization could trade generalizability for performance, and produce over-conservative policies.

For this reason, training with physical robots remains a viable alternative, and this work will therefore focus on real-world learning, where the agent will learn only from real-world interactions, bypassing any type of simulator and the issues that might arise with it.
The review by \citet{annurev2024DRLrealworld} highlight that real-world learning introduces mainly two new challenges, one of them being how to accelerate training with real-world samples, which are clearly limited in number when compared to simulators that can simulate thousands of environments in parallel \cite{rudin2021learning}.
This paper tries to tackle the sample efficiency limitation by applying continual learning \cite{abel_definition_2023} to the \gls{ar} challenge. Specifically, we consider the setup where a set of track layouts can be available in a pre-training phase, and an unknown track, with possibly a different floor, will make up the test condition. 

A first crucial branch of related work analyzes sample-efficient \gls{rl} algorithms. 
\gls{sac} \cite{haarnoja2018sac}, one of the most diffused off-policy algorithms,  constitutes the core component of this paper. While different extensions improving the sample-efficiency of the original algorithm exist (e.g. \cite{bhatt2024crossq, lee2025simba}), we preferred to base our work on the more widespread open-source implementation of \gls{sac} \cite{stable-baselines3}, which has been extended successfully with custom features for real-world learning (asynchronous architecture \cite{async_sac}, generalized \gls{sde} \cite{raffin2021smooth}), and leave the experimentation with novel algorithms to future work.
Similarly, model-based methods have had great success in terms of sample complexity and have been successfully deployed on physical robots \cite{rothfuss2024simfsvgd, brunnbauer2022latent}. However, the added training compute can outweigh the benefits when computational resources are tight.

Real-world training further makes it highly impractical to present the learner with substantially different environments and hence the agent usually has to learn sequentially, i.e. by possibly having access to a specific environment only once. 
Continual learning offers the tools to make this possible.

\subsection{Continual Learning and \acrlong{rl}}
Continual \gls{rl} addresses the degrading performance of neural networks under changing tasks or state distributions \cite{abel_definition_2023}.
Some of its typical challenges are catastrophic forgetting (reduced performance after task switch) and loss of plasticity (inability to adapt after a task switch).
Typical strategies to avoid forgetting include replay buffer management, regularization, or network segregation \cite{crl_survey}. CLEAR \cite{clear}, for instance, combines replay buffer reuse and regularization and shows good performance on the CORA benchmark \cite{cora}, while methods such as \gls{ere} \cite{ere} emphasize newer samples to accelerate adaptation.
Focusing on loss of plasticity instead, different metrics have been developed \cite{abbas_loss_2023, cbp2024}.
Plasticity has been studied through metrics like dormant neurons (dead units) \cite{redo} or average weight magnitudes \cite{l2_init, cbp2024, abbas_loss_2023}, and mitigated by reinitializing biased network heads \cite{gran_turismo_overtaking} or low-utility units, as with \gls{cbp} \cite{cbp2024} and in \gls{redo} \cite{redo}.
Many continual learning methods have been proposed and show increasing performance in simulation. However, few ideas are tested on a physical system.
This work will focus on the racing setting within the RoboRacer format, and particularly on loss of plasticity.

\subsection{\acrlong{ar} and Roboracer}
The Roboracer \gls{ar} competition \cite{pmlr-v123-o-kelly20a-f1tenth} (formerly F1TENTH) provides a key robotic challenge, as participants have to adapt to unseen track layouts under strict time constraints, demanding efficient generalization, either by zero-shot transfer, rapid fine-tuning, or leveraging pre-trained policies.
Classical methods for \gls{ar} such as \gls{pp} \cite{purepursuit}, \gls{mpcc} \cite{mpc_implementation_basis}, and \gls{map} \cite{map} are the most performant on the RoboRacer platform \cite{evans_unifying_2024}, and are also widespread across different \gls{ar} competition formats (e.g. in \cite{betz2023tumindy, raji2024erautopilot}).
\gls{rl} methods, however, provide an interesting alternative, as they achieved \gls{sota} performance in simulated car racing \cite{wurman_outracing_2022} and real-world drone racing \cite{drone_racing}.

On the RoboRacer physical platform, different \gls{rl} methods have been explored without clearly demonstrating high-performance: domain randomization \cite{e2e_lidar_driven_rl}, direct sim-to-real deployment \cite{train_in_austria, brunnbauer2022latent}, and image pretraining with online finetuning (FastRLap) \cite{fastrlap}.
Furthermore, \citet{hildisch2025drive} show that, with a residual policy structure, an \gls{rl} controller can drive faster than a classical controller, and can learn significantly faster than an end-to-end controller, reaching a higher performance level than classical control methods for \gls{ar}.

This work addresses the main issues of \cite{hildisch2025drive} in generalization and presents a continual learning framework based on \gls{cbp} that is able to learn on multiple tracks, generalize to unseen tracks, and continually adapt to a new layout, without using a residual policy structure.
Furthermore, offline \gls{rl} offers an alternative \cite{offlineRL_on_real_robot}, enabling pre-training on a dataset of diverse policies, and even outperforming these later in some cases \cite{OfflinetoOnlineRL}.
This work compares the proposed continual learning framework to an offline framework based on \gls{iql} \cite{kostrikov_offline_2021}.

\section{METHODOLOGY}

\subsection{Evaluation Setup}
\label{ssec:evaluation_setup}
\textbf{Physical System -} For evaluation, we replicate the platform of \cite{baumann_forzaeth_2024} and \cite{hildisch2025drive}, extending the training framework of the latter.
Training is conducted on a remote computer with a NVIDIA GeForce RTX 4070 laptop GPU. Note that training directly on the car was also successful, but the increased computational load induced higher variance in the actor's control frequency and reduced the battery lifetime, making training more time-consuming.

\textbf{Human-Monitored Training -} The safety filter used in \cite{hildisch2025drive} decides based on the relative heading to the intended race line whether to use the \gls{rl} controller or the baseline controller.
This safety filter does not recognize if the car is about to crash when the car is sliding out in corners but still stays parallel to the race line, which can easily happen at high speeds. Furthermore, tracking progress on a residual \gls{rl} approach can be less clear, due to the interplay between the \gls{rl} agent and the baseline controller.
For these reasons, we chose to not use the residual \gls{rl} structure, but to follow the usual structure where full agency is relinquished to the \gls{rl} agent, following what \cite{hildisch2025drive} names an end-to-end controller.
Although the method requires some initial human intervention, continual strategies improve sample efficiency.
As a result, the agent can learn a basic slow policy after only a few episodes, and after a few minutes, only sporadic operator interventions are needed.

\textbf{Train for Generalization, Fine-tune for Racing -} We train our setup on 4 different tracks with 2 different tire-floor combinations (see \Cref{fig:track_setup} for details) for a total of 100'000 gradient steps, which corresponds to 40-60 minutes of training. 
On the final track, we only train for performance.
Practically, this means we keep a maximum of 200'000 samples in the buffer during training and prune overflowing samples at the end of training according to a diversity measure applied trajectory-wise. However, in our tests, this number was only reached after the last training track and therefore will only have an influence for further training.
Such a diversity method consists of a dimensionality reduction of the concatenated observation and action vector by PCA, where we keep the 10 main principal components accounting for 95\% of the variability. This is followed by \gls{knn} with k=5. The per sample diversity score is calculated as the average distance $s_i$ to the k nearest neighbors. From this, we get the final sample priority as $P(i)=1/rank(s)$, where the rank is the sorted position in the replay buffer. We define the trajectory priority as the average score of all samples of the trajectory and keep all samples from the trajectories with the highest priorities. This was inspired by previous approaches that tried to tackle the continual learning problem with buffer management strategies \cite{crl_survey}.
Keeping older samples during training for generalization prevents forgetting. On the final race track, we start with 5'000 samples from the previous replay buffer and let the agent adapt more, accepting some forgetting. The samples are again obtained by employing the diversity filter method. First experiments with an entirely fresh replay buffer were unsuccessful: after a few moments of training the performance dropped severely and the car started acting as in early training.

\subsection{\acrlong{sac}}
Our setup builds on top of the \gls{sac} \cite{haarnoja2018sac} implementation from \gls{sb3} \cite{stable-baselines3}.
We use Async-SAC-1 from \cite{async_sac} implemented as two separate \acrshort{ros} nodes: an acting node and a learning node. The acting node runs at 20 Hz, while the learning node runs without time constraints.
In practice, our setup ran with an \gls{utd} ratio between 1 and 1.7, which was primarily influenced by the sampling time for the increasing replay buffer size.
Furthermore, as previous work has highlighted \cite{bhatt2024crossq}, a higher \gls{utd} ratio would not have been beneficial for the standard \gls{sac} algorithm.

To improve exploration efficiency, we use \gls{gsde} \cite{raffin2021smooth} from the \gls{sb3} implementation.
The most important hyperparameters are summarized in \autoref{tab:sac_hyperparameters}. This setup, as well as observation space, action space, and reward function, is shared between all considered methods.
Trainings were repeated for three different random initializations of the network in the case of real-world experiments, and five different random initializations for the simulated experiments.

\begin{table}[h]
    \caption{Hyperparameters for the asynchronous SAC implementation.}
    \label{tab:sac_hyperparameters}
    \centering
    \begin{tabular}{l|c}
    \hline
    \textbf{Hyperparameter}     & \textbf{Value} \\ \hline \hline
    Network dimension           & 2x256 \\ \hline
    Activation function         & ReLU \\ \hline
    Mini-batch size             & 128 \\ \hline
    Learning rate               & 0.003 \\ \hline
    \end{tabular}
    \hfill
    \begin{tabular}{l|c}
    \hline
    \textbf{Hyperparameter}     & \textbf{Value} \\ \hline \hline
    Discount factor             & 0.99 \\ \hline
    \gls{gsde} sampling rate          & 20 steps \\ \hline
    n-step TD                   & 3 steps \\ \hline
    \end{tabular}
\end{table}

\textbf{Observation Space -} The state information used in the observation space is acquired from the sensor data following \cite{baumann_forzaeth_2024}. The components and the corresponding dimensions are listed in \autoref{tab:observation_space}. The tire forces are estimated using the Pacejka tire model and give the car some information about the grip on a specific floor. The parameters for the model are estimated before training on a new floor through system identification according to \cite{dikici2025learningbased}. The observations are flattened before they are passed as inputs to the networks. Therefore, we define the observations as $\bm{o_t} = [\Delta t,c,s_t,s_{t-s},L,d,\theta,\bm{v},\dot{\psi},\bm{F},\alpha_f, \alpha_r,\bm{p}_{race},\bm{p}_{in},\bm{p}_{out}]^T \in \mathcal{O} \subset \mathbb{R}^{138}$, with full definition of the symbols available in \Cref{tab:observation_space}.

\begin{table}[h]
\caption{Observation space.}
\label{tab:observation_space}
\centering
\resizebox{\columnwidth}{!}{%
    \begin{tabular}{c|l|c|c}
        \hline
        \textbf{Symbol}             &   \textbf{Observation}       & \textbf{Range}  &   \textbf{Dimension} \\ \hline \hline
        $\Delta t$     &    Time between observations           & $(0,0.1]$ s  &   1 \\ \hline
        $c$            &    Collision                           & $\{0,1\}$  &   1 \\ \hline
        $s_t$            &    Longitudinal frenet coordinate      & $[0,120] m$  &   1 \\ \hline
        $s_{t-1}$        &    Previous $s$                        & $[0,120] m$ &   1 \\ \hline
        $L$            &    Length of current track             & $[0,120] m$  &   1 \\ \hline
        $d$            &    \makecell[l]{Lateral deviation \\ from proposed race line}          & $[-2,2] m$  &   1 \\ \hline
        $\theta$       &    \makecell[l]{Relative heading to \\ the proposed race line}      & $[-\pi,\pi]$  &   1 \\ \hline
        $\bm{v}$        &    \makecell[l]{Longitudinal and lateral \\ velocity in car frame}        & $[v_{min},v_{max}]$  &   2 \\ \hline
        $\dot{\psi}$            &    Yaw rate            & $[-13,13] °/s$  &   1 \\ \hline
        $\bm{F}$            &    Front and rear tire forces (x,y,z)   & [$-40,40] m$  &   2x3 \\ \hline
        $\alpha_f, \alpha_r$            &    Front and rear tire slip angles   & $[-\pi,\pi] m$  &   2 \\ \hline
        $\bm{p}_{race}$     &    \makecell[l]{Points along proposed \\ race line in car frame}        & $[-136,136] m$  &   2x20 \\ \hline
        $\bm{p}_{in}$       &    Lateral velocity in car frame        & $[-136,136] m$  &   2x20 \\ \hline
        $\bm{p}_{out}$      &    Lateral velocity in car frame        & $[-136,136] m$  &   2x20 \\ \hline
        
    \end{tabular}}
\end{table}

\textbf{Action Space -} Actions at time step $t$ are defined as $\bm{a_t} \in \mathcal{A} = [-\delta_{\max}, \delta_{\max}] \times [v_{\min}, v_{\max}] \subset \mathbb{R}^2$, where $\delta_{max}=0.42$ is the steering angle limit and $v_{\min}=0.5$ and $v_{\max}=10$ are the velocity limits. The positive lower velocity limit was crucial for fast early learning.

\textbf{Reward Function -} The reward function consists of three parts. The time normalized stepwise advancement reward $r_{sa}=\frac{s_t - s_{t-s}}{v_{\max} \cdot \Delta t}$ and the collision penalty $r_{collision}=-\mathbf{1}_{collision==True}$ are used in simulation and on-car. For the human-monitored on-car learning, we add another penalty $r_{drive}=-\mathbf{1}_{drive-button==False}$ for releasing the safety drive button which has to be pressed by the human operator while the car autonomy is active. The partial rewards are multiplied by a weighting factor, which results in $r_{total}=w_{sa} \cdot r_{sa} + w_{collision} \cdot r_{collision} + \mathbf{1}_{on-car} \cdot w_{drive} \cdot r_{drive}$. We empirically chose the weights to be $w_{sa}=20$, $w_{collision}=16$, and $w_{drive}=10$. 

\subsection{Continual Learning extensions for \gls{sac}}
\label{ssec:CBP-SAC}
Our pipeline integrates \gls{cbp} \cite{cbp2024} into the \gls{sac} algorithm. In particular, we apply it to all layers of the actor and the critic networks. The original implementation of \gls{cbp} only includes one-to-one layer transitions. However, the \gls{sac} actor network has 2 heads for mean and \gls{sd}. We extend the instantaneous utility function as follows with a summation over the output layers:

\[
y_{l,i,t} \;=\;
\frac{\bigl|h_{l,i,t} - \hat{f}_{l,i,t}\bigr|}
{\displaystyle \sum_{j=1}^{n_{l-1}} \bigl|w_{l-1,j,i,t}\bigr|}
\;\cdot\;
\sum_{m=1}^{H_{l+1}}
\;\sum_{k=1}^{n_{l+1}}
\bigl|\,w^{(m)}_{l,i,k,t}\bigr|
\]

With $H_{l+1}$ being the cardinality of similarly shaped output layers. $h_{l,i,t}$ is the current activation of $i$ in layer $l$, $\hat{f}_{l,i,t}$ is the running average of the activation and their difference is the mean-corrected activation. $n_{l-1}$ and $n_{l+1}$ are the number of units in the previous and next layer respectively. $w^{(m)}_{l,i,k,t}$ is the units k-th outgoing weight of the head with index $m$ and  $w_{l-1,j,i,t}$ the units j-th incoming weight.

We combine \gls{cbp} with L2 Init \cite{l2_init}, a regularization-based continual learning method, that adds an L2-loss term between the current and the initial network weights to the learning objective. This is applied to the actor and the critic networks and prevents the network weights from growing indefinitely, which is a common phenomenon that induces loss of plasticity.

\begin{figure}[h]
    \centering
    \includegraphics[width=\columnwidth]{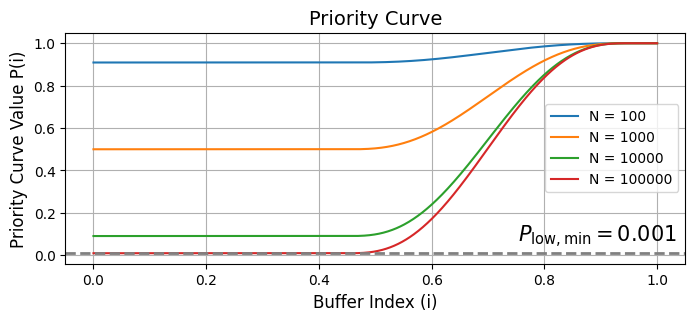}
    \caption{Priority curve for sampling from the replay buffer with emphasis on recent experiences. Buffer index (i) represents relative positioning in the buffer, where i~$=1$ indicates the newest sample and i $=0$ indicates the oldest sample.}
    \label{fig:aere_n}
\end{figure}

Lastly, we employ a variation of the \gls{ere} \cite{ere} sampling strategy, which assumes a fixed size replay buffer and an alternation between a data collection phase and a parameter update phase. In the latter, the training experiences are uniformly sampled from a stepwise shrinking replay buffer. Because we use the asynchronous \gls{sac} setup with a continuously growing replay buffer, the algorithm needs to be adapted.
Instead of shrinking the replay buffer, we define a priority curve (\Cref{fig:aere_n}) which is evaluated for the current replay buffer occupation during sampling. The curve has three parts, a high priority part for new samples with priority $P_{high}(i)=1.0$, an adaptive low priority part, and a transition spline curve in between which follows the quintic smooth step function $P_{transient}(i)=6 \cdot t^5 - 15 \cdot t^4 + 10 \cdot t^3, i \in [c_{low},c_{high}]$, with $c_{low}$ and $c_{high}$ being the percentages of low and high priority samples respectively. We let the lower priority be dependent on the buffer occupation according to $P_{low}(i) = 1 / (1 + \eta \cdot N) > P_{low,min}$, where $\eta$ is a decay factor and $N$ is the replay buffer occupation. This has the effect that the priority for older samples shrinks with increasing buffer occupation. For all experiments, we set those values to $c_{low}=0.5$, $c_{high}=0.95$, $\eta=0.001$, and $P_{low,min}=0.001$. This method gives newer samples more weight, allowing the policy to adapt to the current state distribution while being reminded of older experiences. Note that recomputing the priority curve and the sampling probabilities can dominate computation time if the replay buffer increases too much. In this case, a solution could be to recompute the probabilities only every n steps.

\subsection{Pre-Training with \acrlong{iql}}
\label{ssec:IQL-CBP-SAC}
As the goal during pre-training is to obtain a policy able to generalize across tracks, we tested an offline \gls{rl} method, able to train from different trajectories previously obtained by training with CBP-FS (\Cref{ssec:sim_experiments}) on the training tracks. 
A subset of trajectories from all the training sessions, including some from early learning, were kept for the pre-training.
During the offline pre-training, we sample uniformly from all experiences instead of relying on the most recent one. We use the \gls{iql} algorithm from \cite{kostrikov_offline_2021}  without modifications and set the expectile for the value loss to 0.7 and the temperature for the advantage weighted behavior cloning weights to 0.1.
Furthermore, we use the same network extensions as for the continual variant, \gls{cbp} and L2 Init, and apply them during pre-training and finetuning.

\section{RESULTS}
\subsection{Simulation Experiments}
\label{ssec:sim_experiments}
To more thoroughly analyze the different algorithms, an initial experiment was run in simulation. 
The ROS simulator described by \cite{pmlr-v123-o-kelly20a-f1tenth} was used, adapted with a custom Pacejka Tire model, identified from real data following \cite{dikici2025learningbased}.
While the model was adapted to closely fit reality, none of the simulation trained models were successfully deployed on the physical platform. 
The map layout was directly transferred from the one used in the real world.

Five different models were compared in this experiment: a baseline implementation from the end-to-end model of \cite{hildisch2025drive} trained from scratch on every map, named \textbf{Baseline-FS} (for \textbf{F}rom \textbf{S}cratch); a continually trained version of the baseline, named \textbf{Baseline-C-$n$}, with $n$ indicating the number of tracks it was trained on; a version of the continual method trained from scratch on every track, named \textbf{CBP-FS}; a version trained continually, \textbf{CBP-$n$}; the offline method, \textbf{IQL-$n$}.
Pre-training followed the same procedure as described in \Cref{fig:track_setup} and in the same order.

\begin{figure}[ht]
    \centering
    \includegraphics[width=\columnwidth]{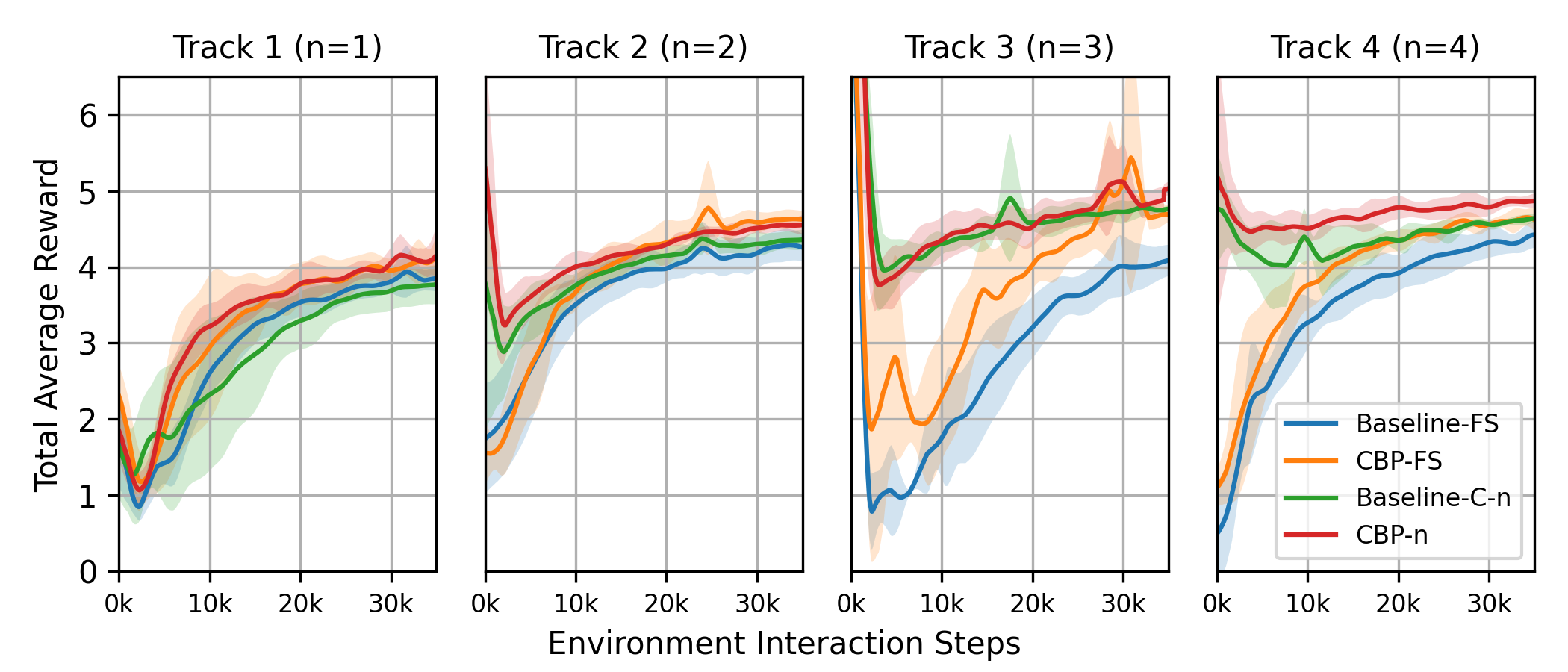}
    \caption{Average total rewards of training in simulation. All configurations are averaged over 5 runs. Shaded areas indicate one \gls{sd} distance.}
    \label{fig:sim_reward_curves}
\end{figure}
\begin{figure}[ht]
    \centering
    \includegraphics[width=\columnwidth]{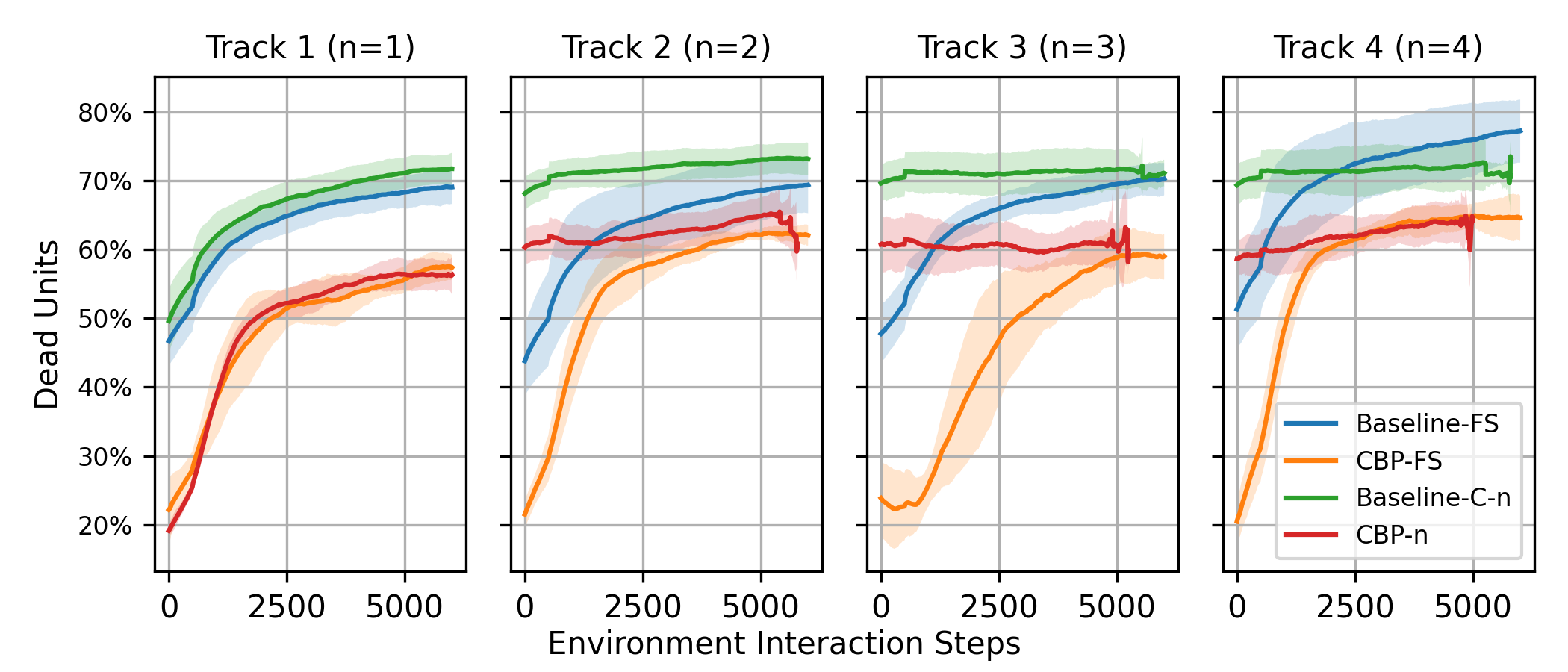}
    \caption{Percentage of dead units in the actor network during training.}
    \label{fig:dead_units_actor}
\end{figure}

In \Cref{fig:sim_reward_curves} the average reward curves show the different learning patterns for the continual methods (i.e. IQL-$n$ is omitted as it is not trained on one track at a time). 
While training on the first two tracks does not isolate clear, distinct patterns, the switch to a larger track layout and a different floor highlights the limitations of the non-continual method. On Track 4, CPB-4 exhibits a marginally higher average reward than Baseline-C-$n$. Furthermore, CBP-FS shows its ability to train more efficiently compared to the other method from scratch, Baseline-FS.
Baseline-C-$n$ also shows a significantly reduced plasticity: looking at the number of dead units of the actor in \Cref{fig:dead_units_actor}, it can be clearly seen that Baseline-C-$n$ flattens at a higher percentage of dead units, slightly above $70 \%$, while CBP-$n$ stays constantly lower, between $60\%$ and $65\%$.

Furthermore, it is interesting to notice that depending on the track, the methods trained from scratch end at consistently different percentages of dead units: while CBP-FS is better than Baseline-FS on every track, both these methods show a higher count of dead units on Track 4 than on the other tracks. 
This result suggests that training on an arguably more complex track (Track 4) first might be detrimental, as it induces a higher loss of plasticity.
\begin{figure}[h]
    \centering
    \includegraphics[width=\columnwidth]{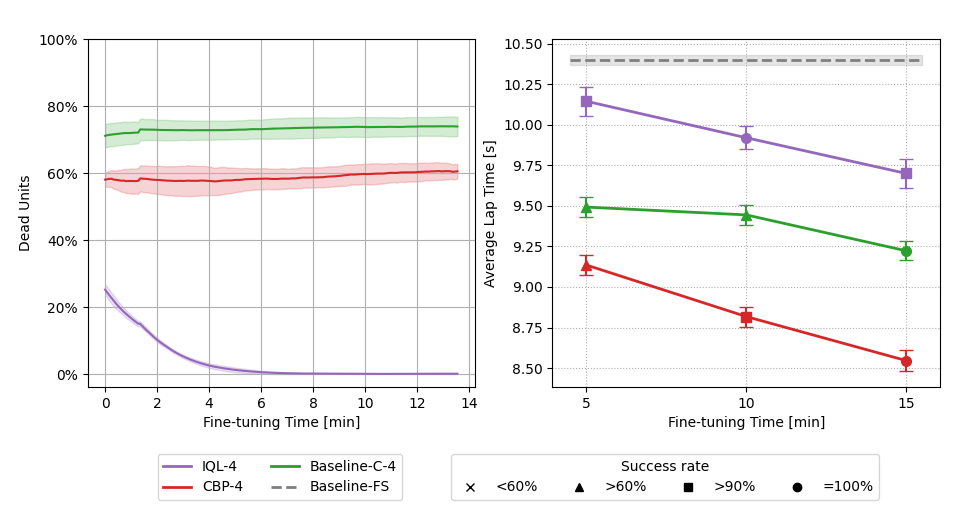}
    \caption{Simulation results on the test track. Left: Percentage of dead units during the finetuning process. Right: Test lap times after 5, 10, and 15 minutes of training on Track 5.}
    \label{fig:sim_lap_time_eval_plot}
\end{figure}
Finally, the different methods are fine-tuned for the same durations as on the physical platform: 5, 10, and 15 minutes. 
As can be seen in \Cref{fig:sim_lap_time_eval_plot}, CBP-n manages to consistently outperform Baseline-C-n, resulting in fewer collisions and always averaging a lower lap time. As expected from the higher amount of dead units shown during pre-training, Baseline-C-n also shows less plasticity. In particular, the lap time reduction is much lower, only a $0.28$s improvement ($9.50$s @ 5 min, $9.22$ @ 15 min, $2.9\%$ relative improvement) for Baseline-C-n, while CBP-n improves by $0.59$s ($9.13$s @ 5 min, $8.54$ @ 15 min, $6.5\%$ relative improvement): for this reason, Baseline-C-n was not further tested on the physical platform.
IQL-n, on the other hand, reports the slowest lap time at all three testing points, but manages to do so with a safer policy at the earlier stages.

Analyzing the plasticity of the actors via the percentage of active units, the trend remains similar to pretraining, as both the continually trained methods face a relatively moderate increase in dead units.
The offline \gls{rl} method, instead, clearly shows a different pattern, by converging to full activation ($0\%$ dead units) and thus promising a highly plastic network. However, the final performance gap to CBP-n turns out to be a shortcoming for the fine-tuning phase.


\subsection{On-car experimental setup}
\label{ssec:on_car_experimental_setup}
Similarly to the simulation setup, the models used for the real-world evaluation were pre-trained on the four physical tracks presented in \Cref{fig:track_setup}: the resulting continual policy is named CBP-4, and similarly, the offline pre-trained policy is IQL-4.
Following \Cref{ssec:evaluation_setup}, the two policies are then finetuned for 15 minutes on the test track, yielding the final policies CBP-5 and IQL-4-CBP, respectively.
All \gls{rl} policies were trained for three different random seeds, and performance was evaluated after 5, 10, and 15 minutes.
The track layout and the tire-floor combination were unseen. The friction was 30\% to 39\% lower than the training tracks, significantly impeding adaptation.
Practically, Tracks 1 and 2 corresponded to a resin-coated floor, Tracks 3 and 4 corresponded to asphalt, and Track 5 was on polished concrete. 
Additionally, we trained a reference policy from scratch (CBP-FS) for 100'000 gradient steps ($\sim50$ minutes) and evaluated its final performance.
We also compare our results to the two well-established model-based classical controllers \gls{map} and \gls{mpc}, following the baseline implementation of \cite{hildisch2025drive}. For this comparison, system identification following the procedure of \cite{dikici2025learningbased} was used, and basic tuning was conducted.
All lap times at the end of fine-tuning are reported in \Cref{tab:lap_times}.

\begin{table}[h]
    \caption{Lap time results after 15 minutes of fine-tuning on the test track. $t_{\mu}$ is the average lap time, $t_{\min}$ and $t_{\max}$ the minimum and maximum lap time respectively, and $\sigma$ the standard deviation}
    \label{tab:lap_times}
    \centering
    \begin{tabular}{l c c c c}
        \hline
        \textbf{Test run}   &   \textbf{$t_{\mu}$ [s]}  &   \textbf{$t_{\min}$ [s]} &   \textbf{$t_{\max}$ [s]}   & \textbf{$\sigma$ [s]} \\ \hline \hline
        \textbf{Classical controllers} &  &  &  &  \\
        MPC & 11.50 & 11.29 & 11.68& 0.095 \\
        MAP & 10.97 & 10.75 & 11.12& \textbf{0.092}  \\ \hline
        \textbf{From scratch} & 10.62 & \textbf{9.94} & 11.74 & 0.431 \\ 
        \textbf{Continual \gls{rl}} & \textbf{10.44} & 9.96 & \textbf{10.99} & 0.199 \\
        \textbf{Offline pre-training} & 12.13 & 10.80 & 14.06 & 1.199 \\ \hline
    \end{tabular}
\end{table}


\subsection{On-car results}
\begin{figure}[t]
    \centering
    \includegraphics[width=\columnwidth]{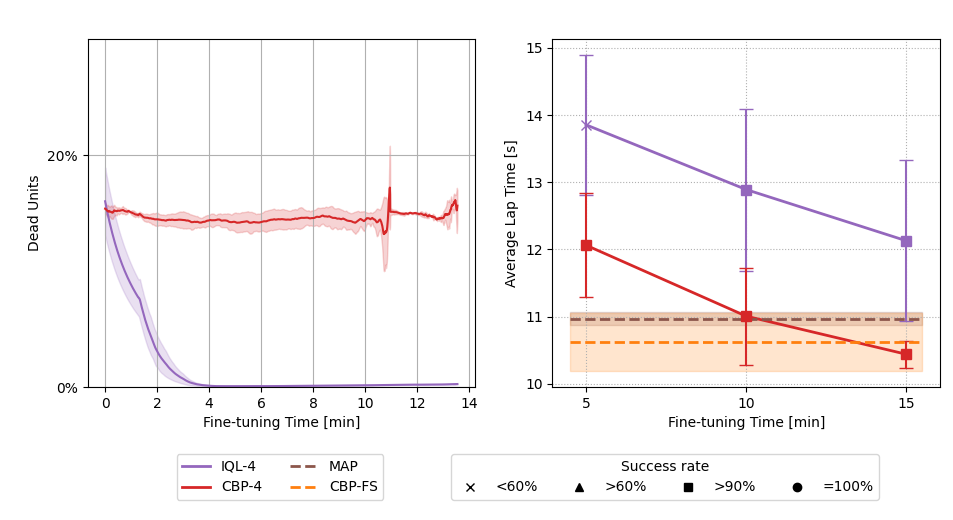}
    \caption{Physical results on the test track. Left: Percentage of dead units during the finetuning process. Right: Test lap times after 5, 10, and 15 minutes of training on Track 5.}
    \label{fig:on_car_combo}
\end{figure}

Lap time evaluations are available in \Cref{fig:on_car_combo}.
After 5 minutes of fine-tuning, one of the three continually learned policies reached \gls{map} performance. After 10 minutes, this was the case on average, and after 15 minutes, all the policies beat the \gls{map} controller. The fastest \gls{rl} policy outperformed the classical baseline by 6.4\%, which is a compatible value with the result achieved by \cite{hildisch2025drive} after 84 minutes.
IQL-4-CBP showed similar results as in simulation, achieving overall a lower performance than the continually trained one, despite its advantage of equal training on all experiences. Dead units metrics, also available in \Cref{fig:on_car_combo}, show a similar pattern as in simulation, particularly for IQL-4-CBP. More interestingly, CBP-5 shows a much lower amount of dead units when compared to simulation (around $18\%$ instead of $60\%$), indicating that the real-world environment requires a much more complex representation.
Almost the same performance, and even the overall lowest lap time, was reached by CBP-FS, albeit with a much longer time at disposal.
\begin{figure}[bh]
    \centering
    \includegraphics[width=0.95\columnwidth]{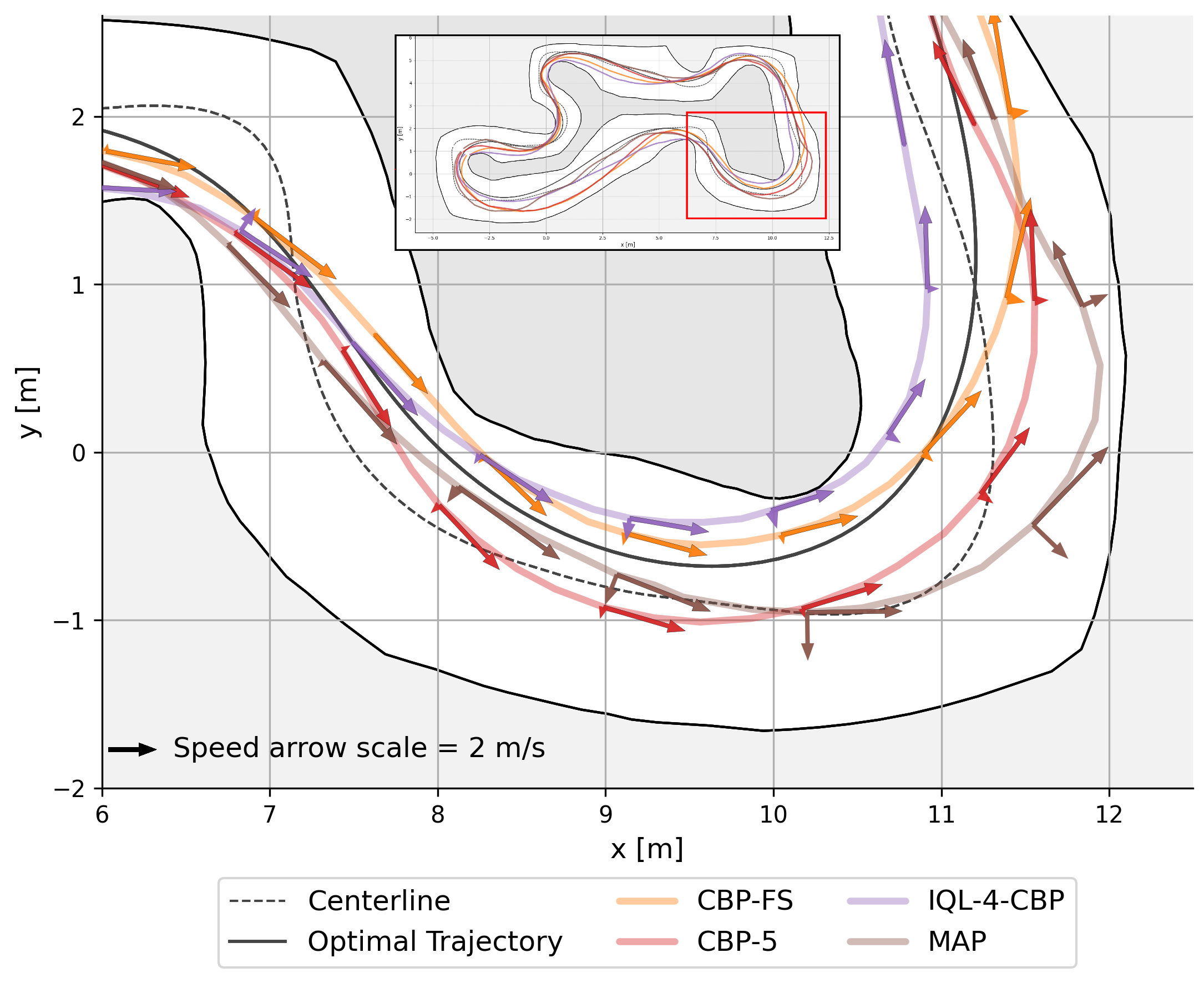}
    \caption{Minimum lap time trajectories on Track 5. Arrows aligned with the trajectories indicate longitudinal velocity, and arrows perpendicular to the trajectory indicate lateral velocity.}
    \label{fig:trajectories}
\end{figure}
The qualitative comparison of the trajectories in \Cref{fig:trajectories} shows that the learned policies that were faster than the \gls{map} controller often chose wider trajectories with lower curvature and managed to carry higher speed in the corners.
CBP-FS, in particular, manages to carry the highest speed through the corner by exiting with a line much closer to the boundaries, possibly indicating a better specialization than CBP-5, which instead tackles the left-hander with a safer trajectory.
Also, \gls{map} seems to commit to a too high velocity before the highlighted corners and needs to overcorrect, induce high slip (notice the higher lateral velocity), and overall exit the corner with a much lower speed than the \gls{rl} policies.
Finally, the offline pre-trained policy chooses a trajectory much closer to the interior boundary in tight curves, which is still possible at lower speeds when the car has not started slipping yet.

\section{CONCLUSIONS}
In this work, we propose a RoboRacer continual \gls{rl} training framework to obtain a generalistic policy able to be finetuned on unseen terrains.
Integrating \gls{sac} with \gls{cbp}, L2 Init, and an adaptation of \gls{ere}, the proposed method manages to be finetuned on a track with lower friction than observed in pre-training, within 10-15 minutes (instead of 82 as previously \cite{hildisch2025drive}), and outperforms previous classical controllers such as \cite{map}.   
Additionally, we explore offline pre-training with \gls{iql}, presenting a policy with promising plasticity, but reduced performance. Our findings concerning network plasticity strengthen the dead units metric proposed in \cite{cbp2024} and further the correlation between it and improved network adaptation capabilities.

Overall, the method presented can be used to train a generalistic policy and to finetune it on a reasonably out-of-distribution task: such an approach could be adapted to other platforms, such as quadruped robots on unseen terrains, dexterous manipulators with unseen materials, etc., where accurate simulation might become impossible and training with real-world samples crucial.
Different limitations could be further investigated, such as the loss of performance induced by offline \gls{rl} pre-training, or the effect of even more sample-efficient \gls{rl} algorithms (e.g. \cite{lee2025simba}).





{\small 
\bibliographystyle{IEEEtranN} 
\bibliography{refs2}
}

\end{document}